\documentclass[dvipsnames,format=sigconf]{acmart}

\setcopyright{none} % Disables the ACM copyright text
\renewcommand\footnotetextcopyrightpermission[1]{} % Removes the conference footer
\usepackage{algorithm,algpseudocode}

\usepackage{makecell}
\usepackage{tikz}
\usepackage{geometry}
\usetikzlibrary{arrows.meta, positioning}

\AtBeginDocument{%
  \providecommand\BibTeX{{%
    \normalfont B\kern-0.5em{\scshape i\kern-0.25em b}\kern-0.8em\TeX}}}

\acmConference[GECCO '26]{The Genetic and Evolutionary Computation Conference 2026}{July 13--17, 2026}{San José, Costa Rica}
\acmYear{2026}
\copyrightyear{2026}
\usepackage{tikz}
\usepackage{pgfplots}
\usepackage{subcaption}
\pgfplotsset{compat=1.18}
\makeatletter
\def\@ACM@checkauthors#1{}
\def\sti@checkauthors#1{}
\makeatother

\begin{document}
\title{Optimizing Appliance Scheduling for Solar Energy Management Using Metaheuristic Algorithms}

\author{Hiba Ahmed, Alexander E.I. Brownlee, Jason Adair, Simon T. Powers}
\email{haa13@stir.ac.uk, alexander.brownlee@stir.ac.uk}
\orcid{1234-5678-9012}
\affiliation{%
  \institution{Computing Science and Mathematics}
  \streetaddress{}
  \city{}
  \state{}
  \country{University of Stirling, UK}
  \postcode{}
}

%\author{Alexander E.I. Brownlee}
%\email{alexander.brownlee@stir.ac.uk}
%\affiliation{%alexander.brownlee@stir.ac.uk
%  \institution{Computing Science and Mathematics}
%  \streetaddress{}
%  \city{Stirling}
%  \country{UK}}

%\author{Jason Adair}
%\affiliation{%
%  \institution{Computing Science and Mathematics}
%  \city{Stirling}
%  \country{UK}
%}

%\author{Simon T. Powers}
%\affiliation{%
% \institution{Computing Science and Mathematics}
% \streetaddress{}
% \city{Stirling}
% \state{}
% \country{UK}}

\renewcommand{\shortauthors}{Trovato and Tobin, et al.}

\begin{abstract}
Renewable energy is essential for meeting future energy demands; however, solar energy generation, which occurs only during daylight hours often does not align with household consumption patterns. Appliances such as cookers, washing machines, and dryers are typically operated according to user preferred schedules rather than solar energy availability, creating a scheduling optimization problem. The objective is to determine optimal appliance start times to maximize renewable energy utilization while minimizing user inconvenience and adhering to system constraints.
This paper presents a metaheuristic approach using Iterated Local Search (ILS) and Simulated Annealing (SA) to optimize appliance start times, while considering appliance operating durations, power consumption, inverter limit, battery state of charge constraints, and solar generation forecasts. Unlike most existing work, the scheduling is extended beyond a single day to accommodate unfinished tasks from previous days (spillover),  ensuring operational continuity and enabling sequential operation across multiple days.
Experimental results show that the sequential multi-day scheduling framework effectively manages system constraints while ensuring user convenience under exclusive solar generation. These findings also open opportunities for future research on multi-objective trade-offs between investment in equipment of various sizes, return on that investment, and user satisfaction.

\end{abstract}



\maketitle
\section{Introduction}
The global shift towards sustainable energy \cite{gielen2019role}
has accelerated the integration of renewable energy sources, particularly solar photovoltaics in residential environments. However, the intermittent and variable nature of renewable energy often does not align with household energy demand. This mismatch occurs in home settings, where appliances such as ovens, washing machines, and dryers are typically operated according to user preferences rather than renewable availability.

To mitigate this mismatch, efficient appliance scheduling is essential for residential energy management systems. The scheduling problem aims to determine optimal appliance start times that maximize renewable energy utilization while minimizing user inconvenience. However, this task is inherently challenging due to its combinatorial nature, involving discrete decision variables, appliance specific operational constraints, inverter capacity limits, and battery charging and discharging restrictions. The complexity increases further in multi-day scheduling scenarios, where appliance operations may span multiple days, creating spillover tasks that must be handled without violating continuity constraints.
Addressing these challenges is critical for enabling effective energy management in smart homes and supporting sustainability goals.

In this paper, we propose a multi-day appliance scheduling model that explicitly accounts for renewable generation variability, household demand, and battery state-of-charge dynamics. The proposed approach employs two metaheuristic techniques Iterated Local Search (ILS) \cite{lourencco2003iterated} and Simulated Annealing (SA) \cite{ kirkpatrick1983optimization} to optimize appliance start times under realistic operational constraints. User dissatisfaction is modeled through deviations from preferred appliance start times, while system constraints include inverter limits, battery charge/discharge capacities, allowable time windows, and spillover management. Simulation results demonstrate the effectiveness of the proposed model in aligning energy demand with renewable supply while minimizing user inconvenience.

\section{Related Work}
Efficient renewable energy management in smart homes is essential for promoting sustainability, lowering energy costs, and enhancing user comfort. Key to this process are Home Energy Management Systems (HEMS), which depend on accurate forecasting and optimized appliance scheduling to fully utilize renewable energy resources while meeting user preferences and operational constraints.

Several comprehensive reviews provide insights into the evolution of optimization techniques for HEMS. Shareef et al. ~\cite{shareef2018review} and Varghese et al. ~\cite{varghese2025optimisation} survey heuristic, metaheuristic, and machine learning based approaches, highlighting objectives such as minimizing electricity costs, reducing the peak-to-average ratio, and lowering carbon emissions. These studies also identify challenges related to scalability, parameter tuning, and real-time implementation in dynamic residential environments.
Heuristic and metaheuristic algorithms have been widely applied to appliance scheduling under renewable energy integration. Imran et al. ~\cite{imran2020heuristic} proposed a hybrid Genetic Algorithm Particle Swarm Optimization approach to schedule household appliances while accounting for renewable generation, battery storage, and user comfort. Kamyab ~\cite{kamyab2025scheduling} applied Simulated Annealing (SA) to price-based HEMS scheduling with user discomfort, but did not consider renewable energy or battery storage; in contrast, the present work integrates PV generation, storage, and multi-day spillover effects. Bastianetto et al. ~\cite{bastianetto2020home} applied SA to household energy scheduling, demonstrating improved performance over benchmark methods; however, their model focuses on single-day scheduling and does not explicitly account for continuous appliance durations or multi-day task spillover. Similarly, Vilar and Affonso ~\cite{vilar2016residential} proposed a residential energy management system using SA to schedule a limited set of shiftable appliances with photovoltaic generation. Their approach respects user time preferences and appliance operation cycles, but primarily aims at reducing grid import costs and does not integrate dynamic battery constraints or multi-day scheduling. ILS is a robust metaheuristic capable of escaping local optima~ \cite{lourencco2003iterated}; however, its application to appliance scheduling with renewable energy integration remains largely unexplored.
Machine learning has also been explored in HEMS to enhance renewable energy integration. Matallanas et al. ~ \cite{matallanas2012neural} developed a neural network–based controller for demand-side management in residential systems integrating PV generation and energy storage, demonstrating improved utilization of local renewable energy and overall system performance. Nakıp et al. \cite{nakip2023renewable} proposed a Forecast Embedded Scheduling framework that integrates a Recurrent Trend Predictive Neural Network to forecast renewable energy availability and schedule household appliances accordingly. Although their method improves forecasting accuracy and scheduling efficiency, it exhibits several limitations under realistic operating conditions. Specifically, the battery model assumes a fixed initial state at the beginning of each day, failing to capture state-of-charge (SoC) continuity and realistic charging/discharging dynamics. Inverter constraint is not explicitly modeled, and appliance scheduling does not fully account for continuous operation active durations, potentially leading to inaccurate energy consumption estimates and suboptimal schedules.

Overall, these studies demonstrate the potential of heuristic, metaheuristic, and machine learning approaches for appliance scheduling. However, most do not fully capture the challenging constraints of isolated microgrid systems, often assuming single day and simplified operational constraints. The joint integration of continuous appliance active durations, dynamic battery-state constraints, user comfort, and multi-day scheduling with task spillover remains underexplored. This gap motivates the development of expanded formulations of the residential renewable energy management problem using only solar energy, without relying on a stable grid connection, and metaheuristic approaches capable of delivering computationally efficient and practically deployable solutions for it.

\section{Problem Description }
The appliance scheduling problem is discrete and combinatorial in nature, as appliance start times must be selected from discrete time slots while respecting appliance active durations, inverter power limits, battery SoC constraints, and the coupling between appliance demand and time-varying PV generation. The interaction of these constraints results in a highly constrained and non-convex search space, making direct or greedy optimization approaches ineffective.

Although the problem is easy to formulate, the solution space is vast. For $N$ appliances, if each appliance can start at any hour between 0 and 23, there are 24 feasible start times per appliance. Consequently, the total number of possible schedules is approximately:
\[
\text{Search Space Size} \approx 24^N
\]
For example:
\begin{itemize}
    \item $3$ appliances: $24^3 = 13{,}824$ possible schedules,
    \item $10$ appliances: $24^{10} \approx 6.3 \times 10^{13}$ possible schedules.
\end{itemize}

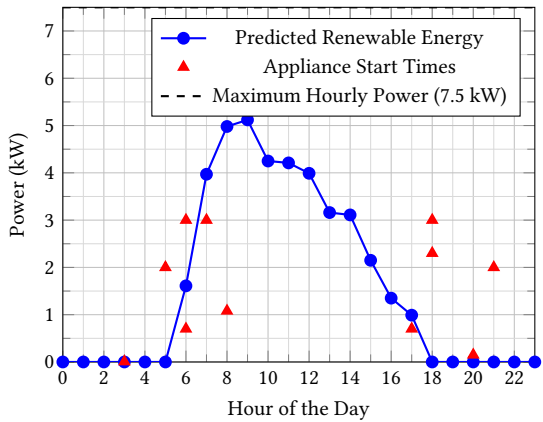
\begin{figure}[b]
\centering
\begin{tikzpicture}
\begin{axis}[
    width=\columnwidth,
    height=0.8\columnwidth,
    xlabel={Hour of the Day},
    ylabel={Power (kW)},
    ymin=0, ymax=7.5,
    xmin=0, xmax=23,
    xtick={0,2,4,6,8,10,12,14,16,18,20,22},
    ytick={0,1,2,3,4,5,6,7},
    legend pos=north east,
    grid=both,
    grid style={line width=.1pt, draw=gray!30},
    major grid style={line width=.2pt,draw=gray!50},
    minor tick num=1,
    tick label style={font=\small},
    label style={font=\small},
    legend style={font=\small},
]

% Predicted renewable energy per hour
\addplot[
    color=blue,
    mark=*,
    thick
] coordinates {
(0,0) (1,0) (2,0) (3,0) (4,0) (5,0) (6,1.61) (7,3.97) (8,4.98) (9,5.12)
(10,4.25) (11,4.21) (12,3.99) (13,3.16) (14,3.11) (15,2.15) (16,1.35)
(17,0.99) (18,0) (19,0) (20,0) (21,0) (22,0) (23,0)
};
\addlegendentry{Predicted Renewable Energy}

% Appliance preferred start times
\addplot[
    only marks,
    mark=triangle*,
    color=red,
    thick
] coordinates {
(5,2) (7,3) (3,0.007) (8,1.08) (20,0.15) (18,2.3) (21,2) (6,0.7) (17,0.7) (6,3) (18,3)
};
\addlegendentry{Appliance Start Times}

% Maximum per-hour power line
\addplot[
    color=black,
    dashed,
    thick
] coordinates {(0,7.5) (23,7.5)};
\addlegendentry{Maximum Hourly Power (7.5 kW)}
\end{axis}
\end{tikzpicture}
\caption{Predicted renewable energy and appliance start times under power constraints for Day 1.}
\label{fig:energy_vs_start_power}
\end{figure}

The practical constraints of the system further increase the difficulty of the scheduling problem. Each appliance has an active duration, preferred start time, and power consumption, while renewable generation fluctuates significantly throughout the day (e.g., from 0~kW at night to over 5~kW at midday). The scheduler must simultaneously satisfy multiple constraints: the 7.5~kW hourly inverter capacity limit, allowable time windows for appliance start times, continuity of each appliance’s required active duration, and cross-day continuity for appliances whose operation extends past midnight, as illustrated in Fig.~\ref{fig:energy_vs_start_power}. This figure highlights the challenge of scheduling appliances to maximize renewable energy utilization while respecting per-hour power constraints.

These constraints interact in complex ways. Running several appliances concurrently may exceed the inverter capacity, whereas shifting appliances away from preferred start times increases user dissatisfaction.

In this context, we consider two metaheuristic approaches aimed at identifying feasible, near-optimal schedules. SA is chosen for its simplicity, robustness, and ability to escape local minima by probabilistically accepting worse solutions. SA has been  applied to complex scheduling problems such as railway crew rostering \cite{hanafi2014hybrid} and airline crew scheduling \cite{emden-weiner1999best}, highlighting its flexibility and suitability for combinatorial problems. In addition, ILS is well suited for appliance scheduling due to its simplicity, ease of implementation, and strong performance in constrained search spaces. By alternating between local search and strategic perturbation, ILS balances exploration and exploitation while maintaining computational efficiency. ILS has been applied to problems such as the single-machine total weighted tardiness problem \cite{congram2002iterated} and the permutation flow shop problem \cite{stutzle1998ils}.

Accordingly, this paper proposes a scheduling optimization framework based on SA and ILS for managing renewable energy in smart homes.

\subsection{Objective Function}
\label{subsec:objective_function}
The primary objective is to minimize the total user dissatisfaction cost across all appliances and time slots $s \in S$, subject to the system constraints defined in Section~\ref{subsec:constraints} and as formalized in \eqref{eq:objective-function}.

\begin{equation}
\label{eq:objective-function}
\min \sum_{n=1}^{N} \sum_{s=0}^{S} x(n, s)\, c(n, s) 
+ \sum_{n=1}^{N} p \Big(1 - \sum_{s=0}^{S} x(n, s)\Big)
\end{equation}

where:
\begin{itemize}
    \item $x(n, s) \in \{0,1\}$ indicates whether appliance $n$ starts at time slot $s$;
    \item $c(n, s)$ is the user dissatisfaction cost for starting appliance $n$ at time slot $s$  as defined in Equation\eqref{eq:appliance_cost}
    \item $s \in \{0,1,\dots,23\}$
    \item $p = 1000$ is the penalty applied only if appliance $n$ is unscheduled ($\sum_{s} x(n,s) = 0$).
\end{itemize}

\begin{comment}
\begin{equation}
\label{eq:objective-function}
\min \sum_{n=1}^{N} \sum_{s=0}^{23} x^{\ast}(n, s)\, c(n, s) + p
\end{equation}

where:
\begin{itemize}
    \item $x^{\ast}(n, s) \in \{0,1\}$ indicates whether appliance $n$ starts at time slot $s$;
    \item $c(n, s)$ is the user dissatisfaction cost for starting appliance $n$ at time slot $s$;
    \item $p$ is a 1000 penalty for unscheduled appliances.
\end{itemize}
\end{comment}
\begin{figure}
            \centering
        \includegraphics[width=1\columnwidth]{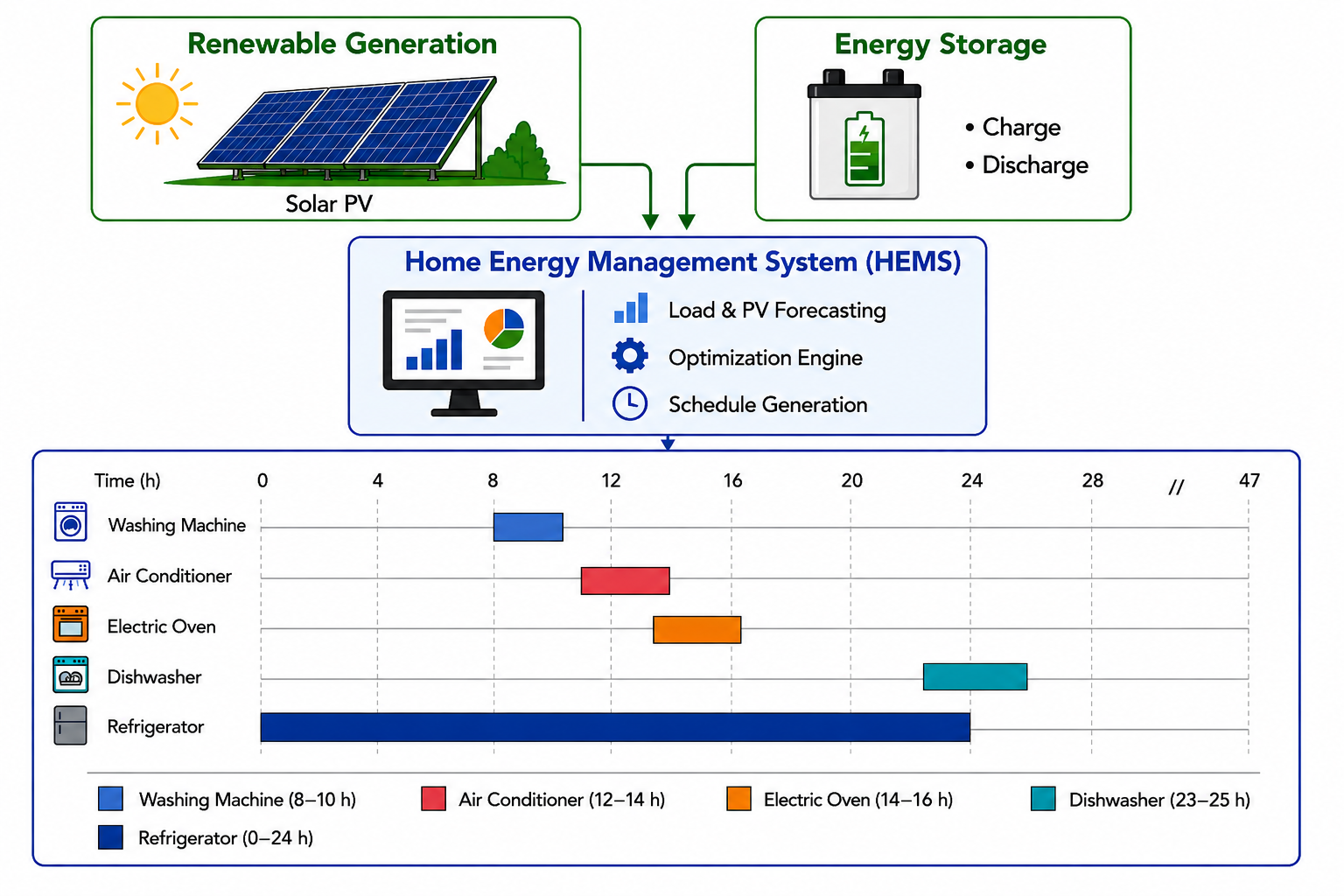}
            \caption{Overview of Home Energy Management System with an example schedule.}
                \label{fig:overview-scheduling-problem}
\end{figure}

% \begin{figure}[htbp]
%     \centering
%     \includegraphics[width=\columnwidth]{Appliance_Cost_Matrix.png}
%     \caption{Appliance Cost.}
%     \label{fig:appliance_cost_heatmap}
% \end{figure}
% \begin{figure}[htbp]
%     \centering
%     \includegraphics[width=\columnwidth]{Appliance_Cost_Matrix.png}
%     \caption{Appliance Cost.}
%     \label{fig:appliance_cost_heatmap}
% \end{figure}

% \begin{figure}[htbp]
%     \centering
%     \includegraphics[width=\linewidth]{Appliance_Cost.png}
%     \caption{Appliance Cost.}
%     \label{fig:appliance_cost}
% \end{figure}

\subsection{User Dissatisfaction Cost}
The cost for each appliance start time depends on how far it is from the preferred start time, with minimal cost near the preferred time and increasing with distance, modeled using a Gaussian distribution, as defined in ~\cite{nakip2023renewable}. This formulation ensures that schedules respect user preferences while allowing flexibility in operation. Mathematically, the dissatisfaction cost for appliance $n$ at time slot $s$ is given by:
\begin{equation}
c(n, s) = 1 - \frac{1}{\sigma_n\sqrt{2\pi} } \exp\left( -\frac{1}{2} \left( \frac{\ s - \mu_n}{\sigma_n} \right)^2 \right)
\label{eq:appliance_cost}
\end{equation}

Where:
\begin{itemize}
    \item $\mu_n$: desired start time of appliance $n$.
    \item $\sigma_n$: represents flexibility in scheduling.
\end{itemize}

Time slots outside the allowed range are assigned a very high cost of 100.

% The cost for each appliance start time is defined as a function of the deviation from the preferred start time. Costs are minimal near the preferred start time and increase with greater deviations, modeled using a Gaussian distribution 

\subsection{Constraints}
\label{subsec:constraints}
1. Operational Constraint

Each appliance must run for its full operational duration once it starts.
\begin{equation}
\label{eq:operational-constraint}
\sum_{s \in \{\, t \in S \mid t \le 2S-(a_n-1) \,\}} x(n, s) = 1, \quad \forall n \in \mathbb{N}.
\end{equation}

Where:
\begin{itemize}
    \item $a_n$: active duration time of appliance $n$.
    \item $x(n,s)$ = 1: indicates appliance $n$ starts at slot  $S$ is the set of all possible time slots (e.g., $S = \{0,1,2,\dots,23\}$) depend on $\sigma_n$.
\end{itemize}

2. Power Consumption Constraint

The total power consumption at any given time slot $s$ should not exceed the inverter capacity:

\begin{equation}
\label{eq:power-constraint}
\sum_{n=1}^{N} P_n(s) \cdot x(n,s) \le \Theta, \quad \forall s = 1, \dots, S
\end{equation}

Where:
\begin{itemize}
    \item $P_n(s)$: power consumption of appliance $n$ at slot $s$ 
    \item $\Theta$: inverter capacity (kW)
\end{itemize}

3.Battery Constraints

The battery state $B(t)$ at time $t$ is updated according to the predicted  generation and power consumption as:

\begin{equation}
\label{eq:battery-update}
B(t) = B(t-1) + G(t) - \sum_{n=1}^{N} P_n(t)\, x(n,t), 
\quad \forall t = 1, \dots, S.
\end{equation}

The battery SoC is constrained to remain within its minimum and maximum allowable limits:

\begin{equation}
\label{eq:battery-limits}
Z_{\min} \le B(t) \le Z_{\max}, 
\quad \forall t = 1, \dots, S.
\end{equation}

The renewable generation available for charging the battery at each time slot is limited by the minimum and maximum bounds, $Q_{\min}$ and $Q_{\max}$ respectively.

\begin{figure}
    \centering
    \includegraphics[width=1\linewidth]{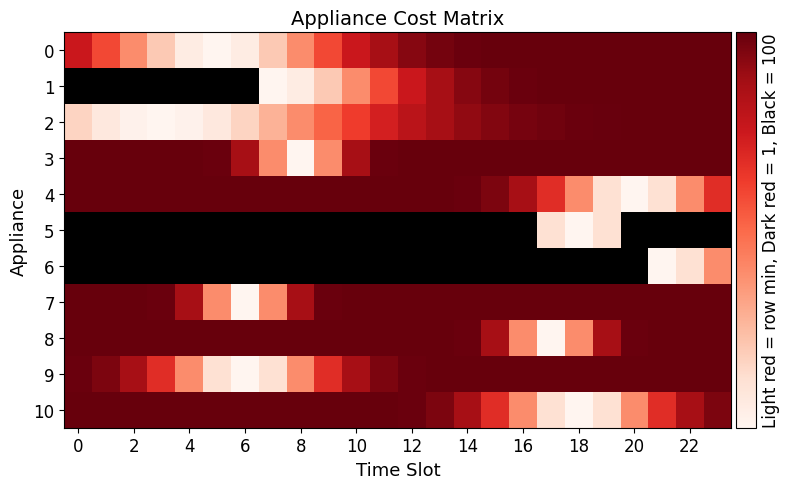}
    \caption{Appliance Cost.}
    \label{fig:appliance-cost-heatmap}
\end{figure}

\begin{comment}
\begin{figure}
    \centering
    \includegraphics[width=1\linewidth]{appliance_cost.png}
     \caption{Appliance Cost.}
    \label{fig:appliance-cost-heatmap}
\end{figure}

\begin{figure}[htbp]
    \includegraphics[width=\columnwidth]{Appliance_Cost_M}
    \caption{Appliance Cost.}
    \label{fig:appliance-cost-heatmap}
\end{figure}
\end{comment}

\section{ METHODOLOGY}

This work considers a smart home energy management system operating without grid interaction, where appliance scheduling is performed using locally available photovoltaic (PV) generation and battery storage. The objective is to schedule appliance start times within user defined operating windows while satisfying all system constraints defined in Equations~\ref{eq:operational-constraint}--\ref{eq:battery-limits} and minimizing user dissatisfaction as formulated in Equation~\ref{eq:objective-function}.
Each appliance is assigned a dissatisfaction cost based on its scheduled start time, as defined in Equation~\ref{eq:appliance_cost}, and visualized in Figure~\ref{fig:appliance-cost-heatmap}.

 Appliance scheduling is performed sequentially using discrete hourly time slots and represented using a binary ON/OFF matrix that captures the activity of each appliance throughout the day. 
\subsection{Schedule Solution Representation}
 \label{sec:schedule-solution-representation}
 candidate schedule is represented as a binary matrix of size $[N, 48]$, where $N$ is the number of appliances and each column corresponds a one hour slot in a 48 hour window (current day + next day),  where columns 0–23 represent the current day, while columns 24–47 represent extended slots for the following day. This extended slots enables spillover modeling for appliances whose operation continues past midnight. A matrix element with value \textbf{1} indicates that the corresponding appliance is \textbf{ON} during that hour, while a value of \textbf{0} indicates it is \textbf{OFF}. The schedule representation is:
 
\[
\mathbf{X} = 
\begin{bmatrix}
x_{1,0} & x_{1,1} & \cdots & x_{1,23} & x_{1,24} & \cdots & x_{1,47} \\
x_{2,0} & x_{2,1} & \cdots & x_{2,23} & x_{2,24} & \cdots & x_{2,47} \\
\vdots & \vdots & \ddots & \vdots & \vdots & \ddots & \vdots \\
x_{N,0} & x_{N,1} & \cdots & x_{N,23} & x_{N,24} & \cdots & x_{N,47} \\
\end{bmatrix}
\]
Where:
\begin{itemize}
    \item Each row represents one appliance.
    \item Columns 0--23: Current day slots.
    \item Columns 24--47: Spillover slots for the next day.
\end{itemize}
This representation allows the scheduler to handle appliance activation durations that span to next day, which is essential for accurate and feasible scheduling.

% where:
% \begin{itemize}
  %   \item Each row represents one appliance.
 %    \item Columns 0–23: current day slots.
   %  \item Columns 24–47: spillover slots for the next day.  
% \end{itemize}

% To illustrate further, for instance consider Appliance 1 with a start time of 3 and an active duration of 2 hours. It operates continuously from 03:00 to 05:00, so columns 3 to 5 in its row are set to 1. If the appliance 3 starts at hour 23 with a duration of 2 hours, it operates from 23:00 to 01:00 the next day. In this case, column 23 in the current day section and column 24 in the spillover section are set to 1, ensuring continuity across midnight and it is carried out to the next day.
\subsection{Sequential Multi-Day Scheduling}

Appliance scheduling is performed sequentially over 86 consecutive days using the proposed metaheuristic optimization techniques, ILS and SA, based on predicted renewable energy data~\cite{nakip2023renewable}. The daily simulation framework, as outlined in Algorithm~\ref{alg:multi_day}, generates multi-day schedules that maximize renewable energy utilization while minimizing user dissatisfaction.
he framework implements a rolling-window approach, in which
system states and inputs including battery SoC and cross-day
appliance spillover are updated daily prior to optimization. This
ensures that appliance activity and battery energy from the previous
day are accounted for when planning the current day’s schedule.

Each day, a 48-hour PV forecast is prepared by combining the current and next day’s predicted generation. The appliance cost matrix is computed, and the metaheuristic optimizer (LS or SA) is invoked to determine the best feasible appliance schedule while respecting system constraints. Deviations between actual and preferred appliance start times are computed, and battery SoC is simulated hour by hour, with infeasible loads corrected to maintain operational feasibility.

The last 24 hours of each optimized schedule are propagated as spillover for the next day, and the end of day battery SoC is carried forward. Daily metrics including fitness values, deviations, and SoC tracking are recorded for subsequent analysis and visualization.
\subsection*{Pseudocode: Multi-Day Simulation Loop}

\begin{algorithm}
\caption{Daily Simulation Framework for ILS and SA}
\label{alg:multi_day}
\begin{algorithmic}[1]
\State \textbf{Initialize:}
\State $SoC \gets 0$ \Comment{Initial battery state-of-charge}
\State $Spillover \gets \mathbf{0}_{N \times 2S}$ \Comment{Cross-day spillover matrix}
\State Allocate storage arrays for fitness and SoC profiles
\For{$d = 1$ to $D$} \Comment{Loop over all days}
    \Statex \vspace{-2mm}
    \State \textbf{Step 1: Prepare 48-hour PV forecast}
    \State $G_{48h} \gets [G_d, G_{d+1}]$
    \Statex \vspace{-2mm}
    \State \textbf{Step 2: Generate cost matrix}
    \State $Cost \gets \text{GenerateCostMatrix}(N, S)$
    \Statex \vspace{-2mm}
    \State \textbf{Step 3: Run metaheuristic optimizer (ILS or SA)}
    \State $(X^*, f_{\text{best}}, f_{\text{init}}, StartTimes, Spillover_{next}) \gets$
    \Statex \hspace{3mm} $\text{Optimizer}(Spillover, SoC, G_{48h}, Cost, \text{constraints})$
    \State Record $f_{\text{init}}$ and $f_{\text{best}}$ for day $d$
    \Statex \vspace{-2mm}
    \State \textbf{Step 4: Compute user-deviation metrics}
    \For{$n = 1$ to $N$}
        \State $Deviation[d][n] \gets StartTimes[n] - DesiredStart[n]$
    \EndFor
    \Statex \vspace{-2mm}
    \State \textbf{Step 5: Simulate battery state for 48 hours}
    \State $SoC_{temp}[0] \gets SoC$
    \For{$t = 1$ to 48}
        \State $G \gets G_{48h}[t]$
        \State $C \gets \sum_{n=1}^{N} X^*[n,t] \cdot P_n$ \Comment{Total appliance load}
        \State $SoC_{temp}[t] \gets \min(Z_{\max}, \max(Z_{\min}, SoC_{temp}[t-1] + G - C))$
        \If{$C > SoC_{temp}[t-1] + G$}
            \State $X^*[:,t] \gets \mathbf{0}$ \Comment{Force feasibility-corrected schedule}
        \EndIf
    \EndFor
    \State $SoC\_profile[d] \gets SoC_{temp}[1:24]$ \Comment{Store first 24 hours}
    \Statex \vspace{-2mm}
    \State \textbf{Step 6: Propagate states to next day}
    \State $Spillover \gets X^*$ \Comment{Last 24 hours become next-day spillover}
    \State $SoC \gets SoC_{temp}[24]$ \Comment{Carry end-of-day battery state}
\EndFor

\Statex \vspace{-2mm}
\State \textbf{Return:} Recorded metrics (fitness, deviations, SoC profiles)
\end{algorithmic}
\end{algorithm}

The daily simulation framework (Algorithm~\ref{alg:multi_day}) generates optimized appliance schedules for SA and ILS, integrating 48-hour PV forecasts, cross-day spillover, battery dynamics, and user-preferences.
The following sections outlined implementation of ILS and SA algorithms to model appliance scheduling in renewable powered  homes.

\subsection{Iterated Local Search (ILS)}
\label{sec:ils}

The ILS framework~\cite{lourencco2003iterated} is implemented to optimize daily appliance start times in a sequential multi-day setting, respecting system constraints (Equations~\ref{eq:operational-constraint}--\ref{eq:battery-limits}) and minimizing user dissatisfaction (Equation~\ref{eq:objective-function}). At the beginning of each day, a feasible initial schedule is generated using the procedure outlined in Algorithm~\ref{alg:initial-schedule}. On the first day, this schedule is created randomly while satisfying all system constraints defined in Equations~\ref{eq:operational-constraint}--\ref{eq:battery-limits}, and is intended to provide a valid starting point for minimizing user dissatisfaction as formulated in Equation~\ref{eq:objective-function}. For subsequent days, spillover from the previous day is incorporated to carry over ongoing appliance activity, while the remaining appliances are randomly assigned feasible start times within their allowed windows. This initialization ensures that each day starts with a schedule consistent with the system state and feasible with respect to all constraints. Once the initial schedule is established, ILS iteratively alternates between two main steps:

\begin{itemize}
    \item \textbf{Local Search (Exploitation):} Small adjustments are made to appliance start times by choosing one appliance uniformly at random, and adding or subtracting a value sampled uniformly at random of between 1--3 hours from the start time (\textit{local shifts} in Table~\ref{tab:tuned_hyperparams}). Candidate schedules are evaluated for feasibility with respect to battery SoC, PV generation, inverter limits, and spillover continuity. Feasible changes that improve the objective function are accepted, updating the current and best schedule. This is allowed to repeat \textit{local max iterations} (Table~\ref{tab:tuned_hyperparams}) without improvement before perturbation.

    \item \textbf{Perturbation (Exploration):} To escape local optima, a random subset of 2 appliances is selected, and larger start-time changes are applied (add or subtract a value chosen unifornly at random from the range 5–8 hours, \textit{perturb shifts} in Table~\ref{tab:tuned_hyperparams}). Candidate schedules are again evaluated for feasibility, and improved solutions are incorporated as the best schedule.
\end{itemize}

This process is repeated for a predefined number of iterations. 
%The procedure is summarized in Algorithm~\ref{alg:ils-spillover-shift}.
\begin{algorithm}
\caption{Initial Schedule Generation}
\label{alg:initial-schedule}
\begin{algorithmic}[1]
\State \textbf{Input:} Appliances, active durations, valid time windows, 
\Statex \hspace{1cm} predicted renewable generation, battery limits, inverter limits, 
\Statex \hspace{1cm} previous-day spillover (if any)
\State \textbf{Output:} Initial schedule matrix, appliance start times

\State Initialize schedule matrix to zeros
\If{previous-day spillover exists}
    \State Copy spillover into current day schedule
\EndIf

\For{each appliance $n$}
    \State Determine feasible start slots within appliance time window
    \State Randomly shuffle feasible slots
    \State $scheduled \gets$ \textbf{False}

    \For{each candidate start slot $t$}
        \State Assign appliance $n$ to slots $t$ to $t + d_n$ in schedule
        
        \State \textbf{Step: Check constraints}
        \If{Battery SoC, inverter, and renewable generation constraints are satisfied over schedule horizon}
            \State Accept start slot $t$ for appliance $n$
            \State Update schedule matrix
            \State $scheduled \gets$ \textbf{True}
            \State \textbf{Break} loop for this appliance
        \EndIf
    \EndFor

    \If{$scheduled = $ \textbf{False}}
        \State Mark appliance $n$ as unscheduled
    \EndIf
\EndFor

\State \textbf{Return:} Schedule matrix and start times
\end{algorithmic}
\end{algorithm}

\subsection{Simulated Annealing (SA)}
\label{sec:sa}

Simulated Annealing (SA)~\cite{kirkpatrick1983optimization} schedules appliances sequentially (Section~\ref{sec:schedule-solution-representation}). A feasible initial schedule is generated using the procedure outlined in Algorithm~\ref{alg:initial-schedule}. At each iteration, one appliance’s start time is adjusted by adding or subtracting a value sampled uniformly at random from the range $[1,6]$ (\textit{Shift range}, Table~\ref{tab:tuned_hyperparams}) to generate a neighboring solution. The neighbour is checked against system constraints (Equations~\ref{eq:operational-constraint}--\ref{eq:battery-limits}) and evaluated using the objective function (Equation~\ref{eq:objective-function}). Improvements are accepted, and worse solutions are probabilistically accepted based on a temperature parameter that decreases linearly. The process tracks the best solution and fitness progression throughout.

\begin{table*}[ht]
  \centering
  \caption{Appliance Information ~\cite{nakip2023renewable}}
  \label{tab:appliance_info}
  \begin{tabular}{lp{1.6cm}ccccc}
    \toprule
    \textbf{Appliance} & \textbf{Power Consumption  (kW)} & \textbf{Duration} & \textbf{Desired Start} & \textbf{Variance} & \textbf{Lower Bound} & \textbf{Upper Bound} \\
    \midrule
    Washing Machine (warm wash)     & 2.000 & 2 & 5  & 3 & 0  & 0 \\
    Dryer (avg. load)               & 3.000 & 2 & 7  & 3 & 7  & 0 \\
    Robot Vacuum Cleaner (charging) & 0.007 & 2 & 3  & 5 & 0  & 0 \\
    Iron                            & 1.080 & 1 & 8  & 1 & 0  & 0 \\
    TV                              & 0.150 & 3 & 20 & 2 & 0  & 0 \\
    Oven                            & 2.300 & 1 & 18 & 2 & 17 & 20 \\
    Dishwasher                      & 2.000 & 2 & 21 & 2 & 21 & 0 \\
    Electric Water Heater 1         & 0.700 & 1 & 6  & 1 & 0  & 0 \\
    Electric Water Heater 2         & 0.700 & 1 & 17 & 1 & 0  & 0 \\
    Central AC 1                    & 3.000 & 2 & 6  & 2 & 0  & 0 \\
    Central AC 2                    & 3.000 & 2 & 18 & 2 & 0  & 0 \\
    \bottomrule
  \end{tabular}
\end{table*}

\section{ Experimental Setup}
\label{sec:experimental_setup}
This section introduces the data, parameters, and constraints used in the household appliance scheduling model, including solar forecast inputs, appliance specifications, and system-level constraints.

\subsubsection*{Solar Generation Forecast}

\begin{figure}
    \centering
\includegraphics[width=1\linewidth]{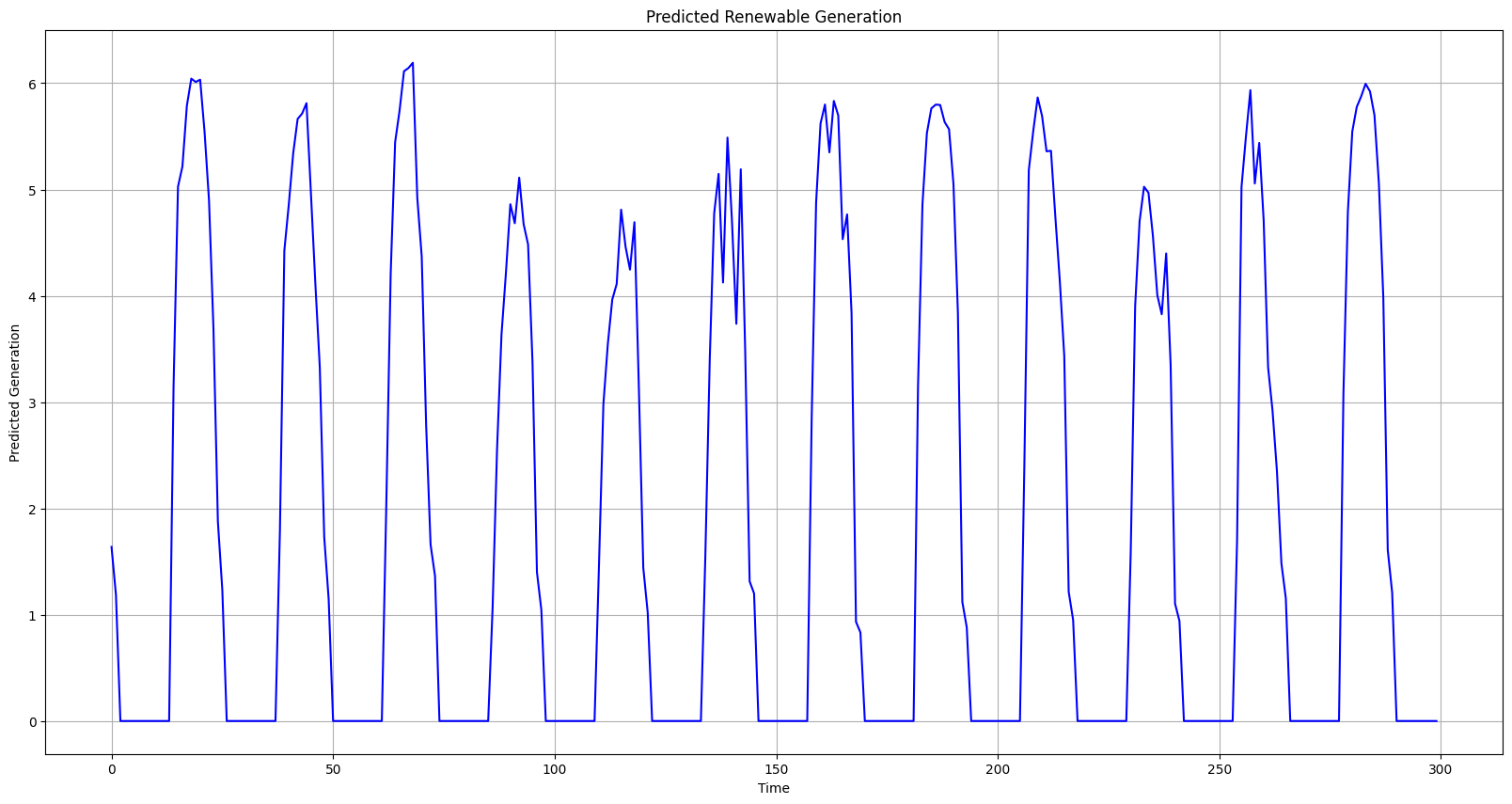}
    \caption{Prediction of solar generation from ~\cite{nakip2023renewable}.}
  \label{fig:solar_forecast}
\end{figure}

Solar PV forecasts for an 86-day period were generated using the rTPNN-FE model~\cite{nakip2023renewable} as in Figure~\ref{fig:solar_forecast}. For each day \( d \in [1,86] \), a 48-hour forecast was defined as:
\[
    G_{48h} = [G_d,\, G_{d+1}],
\]
where each \( G_d \) is a 24-hour PV forecast. These rolling 48-hour predictions serve as inputs to the scheduling algorithm, which operates over \( S = 48 \) one-hour time slots and allows appliance operations to extend into the following day. When scheduling daily appliance usage, the corresponding 48-hour forecast \( G_{48h} \) for day \( d \) is used to align operation with expected solar availability.

\subsubsection*{Appliance Dataset}

This study considers a virtual smart home environment comprising \( N = 11 \) residential appliances, with specifications sourced from a publicly available dataset on GitHub ~\cite{nakip2023renewable}. Each appliance \( n \in \mathcal{N} \) is characterized by the key attributes summarized in Table~\ref{tab:appliance_info} as:

\begin{itemize}
    \item Power consumption per time slot, \( E_n \)
    \item Required continuous active duration, \( a_n \)
    \item User-preferred start time
    \item Variance, \( \sigma_n \), representing the flexibility around the desired start time
    \item Lower and upper bounds defining hard constraints on valid start times
\end{itemize}

These parameters define each appliance within the scheduling model, ensuring all devices run non-preemptively  that is, once started, each appliance must operate continuously for \( a_n \) consecutive time slots.

\subsubsection*{System Constraints:}
The scheduling system must satisfy the operational, power, and battery constraints described in Subsection~\ref{subsec:constraints} (Eqs.~\ref{eq:operational-constraint}–\ref{eq:battery-limits}), with parameter values summarized in Table~\ref{tab:system_constraints}.

\begin{table}[htbp]
  \centering
  \caption{System Constraints}
  \label{tab:system_constraints}
  \begin{tabular}{ll}
    \toprule
    Parameter & Value \\
    \midrule
    $Q_{\min}$ & 0 \\
    $Q_{\max}$ & 50 \\
    $Z_{\min}$ & 0 \\
    $Z_{\max}$ & 100 \\
    $\Theta$ & 7.5  \\
    \bottomrule
  \end{tabular}
\end{table}

By embedding PV forecast data directly into the scheduling process, the system anticipates energy availability and proactively adjusts appliance start times. This integration enables more intelligent and adaptive scheduling decisions that improve energy efficiency and user satisfaction.

\subsubsection*{Hyperparameter Tuning:}
Before the multi-day simulations, grid search was conducted to identify optimal hyperparameters for both algorithms, ensuring fair and effective comparisons. The final selected values for iterated local search (ILS) and simulated annealing (SA) are summarized in Table~\ref{tab:tuned_hyperparams}.

\begin{table}[t]
\centering
\caption{Tuned hyperparameters used in experiments.}
\label{tab:tuned_hyperparams}
\begin{tabular}{lll}
\toprule
Algorithm & Parameter & Value \\
\midrule
SA  & Initial temperature   & 1000 \\
SA  & Cooling rate          & 0.993 \\
SA  & Shift range           & 6 \\
ILS & Perturbation max      & 2 \\
ILS & Perturb shifts        & [8, -5, -8, 5] \\
ILS & Local max iterations  & 10 \\
ILS & Local shifts          & [3, -3, 1, -1] \\
\bottomrule
\end{tabular}
\end{table}

\section{Results}
\begin{figure}
    \centering
    \includegraphics[width=1\linewidth]{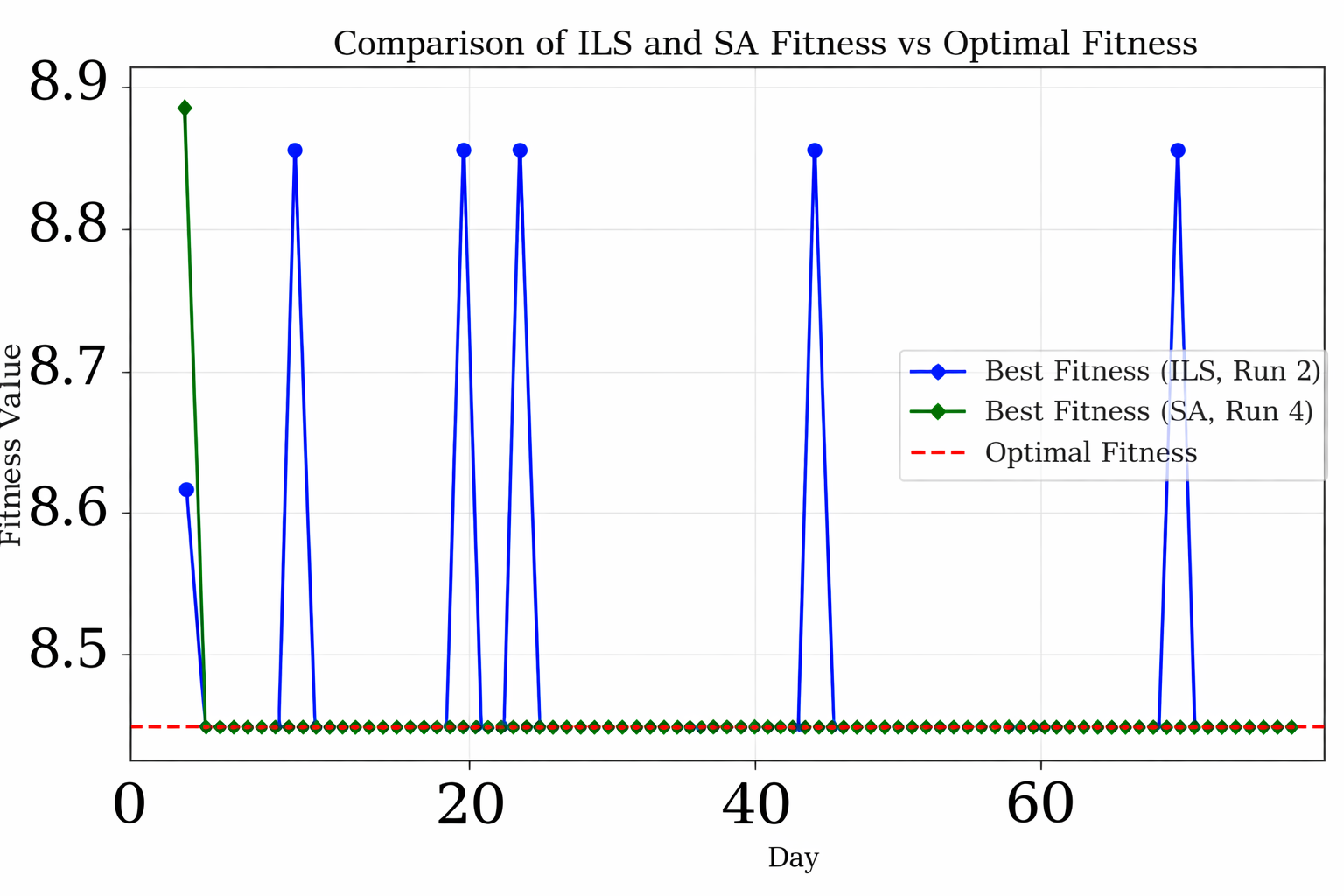}
     \caption{User dissatisfaction: ILS, SA, and optimal.}
    \label{fig:dissatisfaction_comparison}
\end{figure}

This section presents the results of the daily multi-day simulation framework (Algorithm~\ref{alg:multi_day}). Simulated Annealing (SA) and Iterated Local Search (ILS) are applied to the appliances listed in Table~\ref{tab:appliance_info} to minimize user dissatisfaction. Figure~\ref{fig:dissatisfaction_comparison} compares the user dissatisfaction achieved by SA, ILS, and the optimal solution. The theoretical optimal fitness is computed by summing, for each appliance, the minimum dissatisfaction cost across all time slots without taking constraints into consideration. This minimum cost corresponds to the user’s preferred start time for each appliance and represents the lowest achievable total user dissatisfaction. It serves as a benchmark to evaluate how closely SA and ILS approximate the best possible schedule solution. Both metaheuristics achieve results near the optimal, demonstrating the effectiveness of the proposed scheduling framework.

\begin{figure}
    \centering
    \includegraphics[width=1\linewidth]{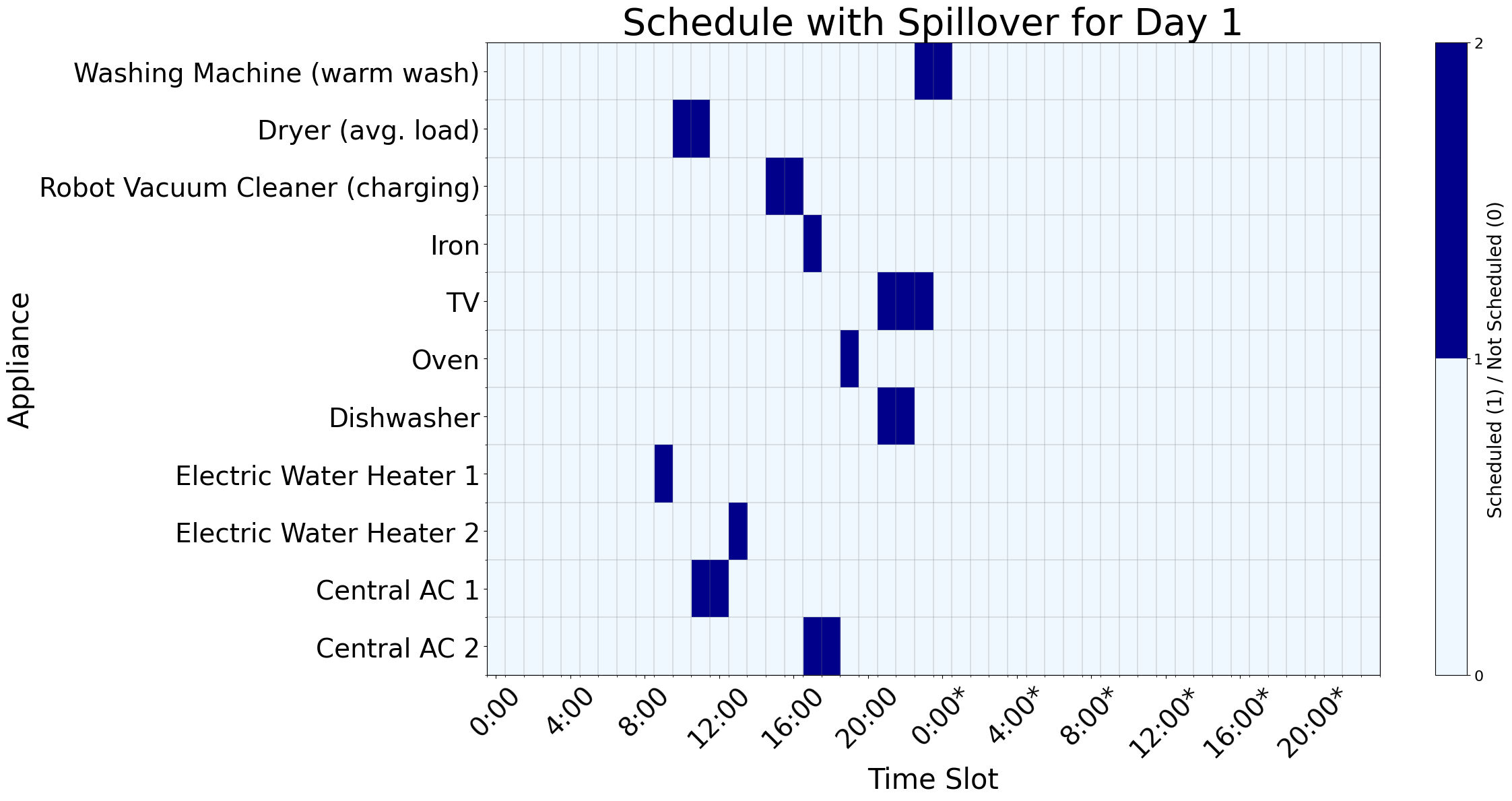}
    \caption{Day 1 schedule (SA) showing washing task spillover into Day 2.}
    \label{fig:day1_schedule}
\end{figure}

\begin{figure}
    \centering
    \includegraphics[width=1\linewidth]{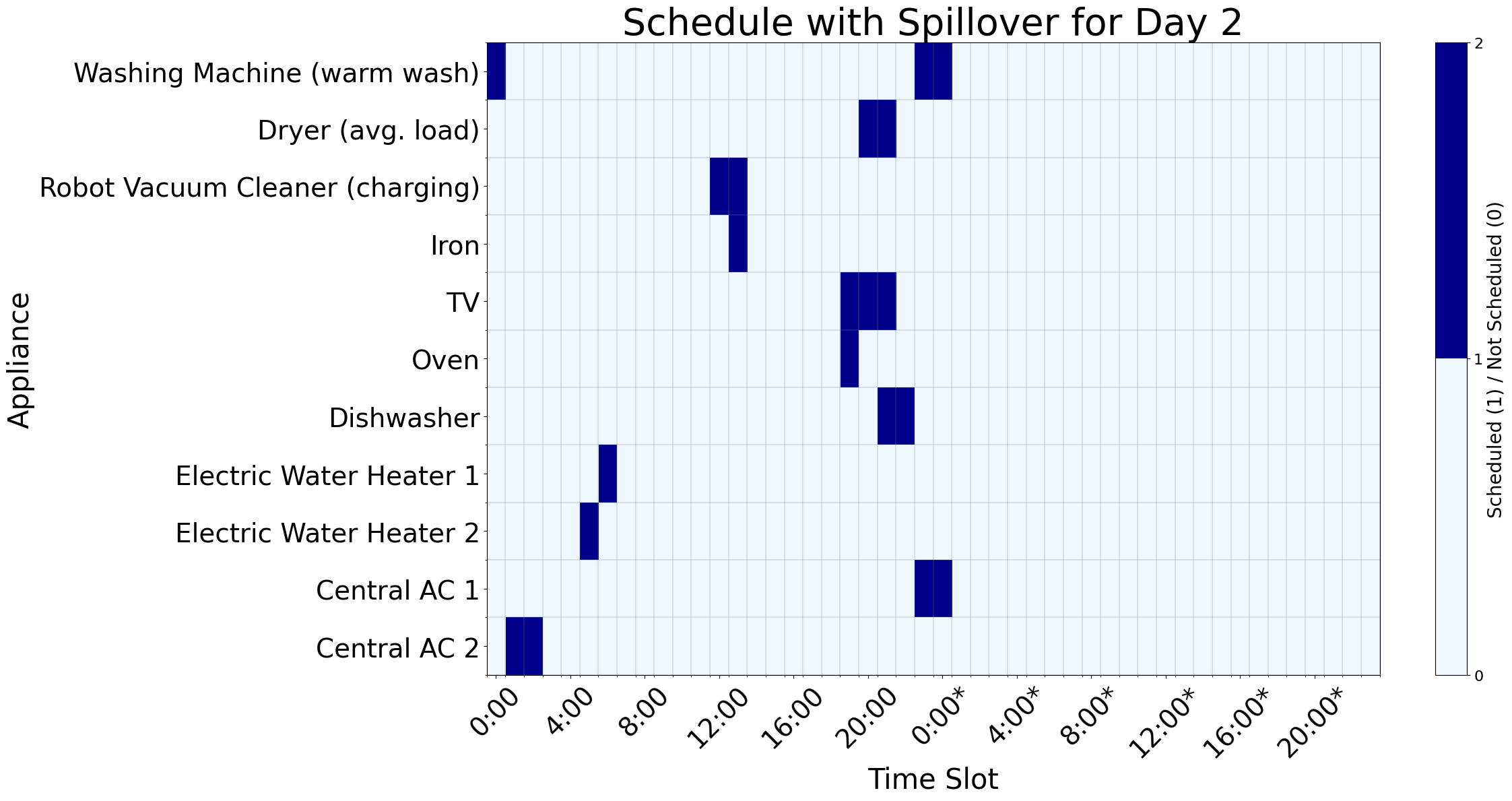}
    \caption{Day 2 schedule (SA) continuing from previous day.}
    \label{fig:day2_schedule}
\end{figure}

\begin{figure}[ht]
    \centering
    % First subfigure
    \begin{subfigure}[b]{0.45\textwidth}
        \centering
        \includegraphics[width=\linewidth]{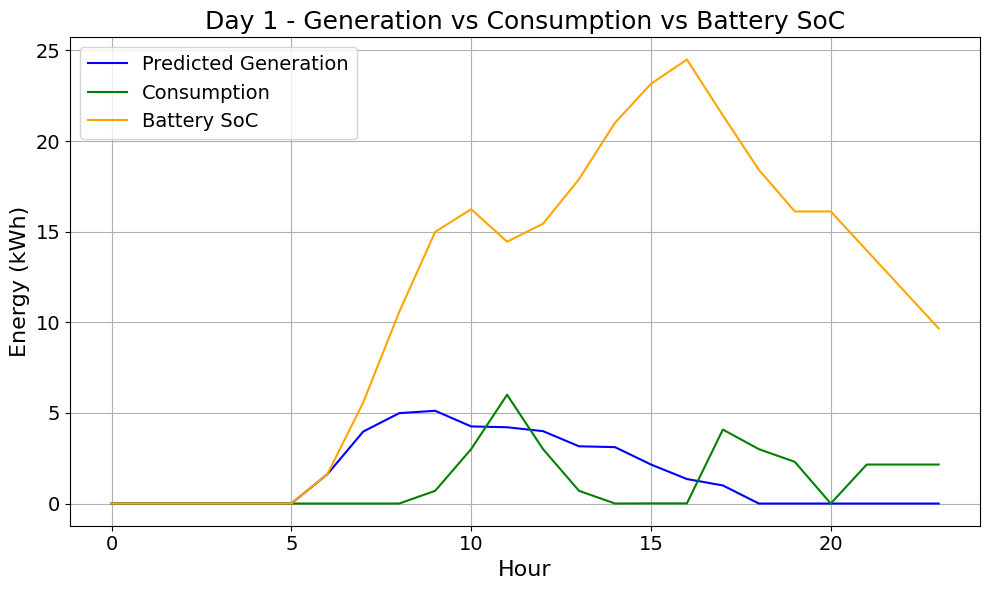}
        \caption{Day 1}
        \label{fig:day1_hourly_tracking}
    \end{subfigure}
    \hfill
    % Second subfigure
    \begin{subfigure}[b]{0.45\textwidth}
        \centering
        \includegraphics[width=\linewidth]{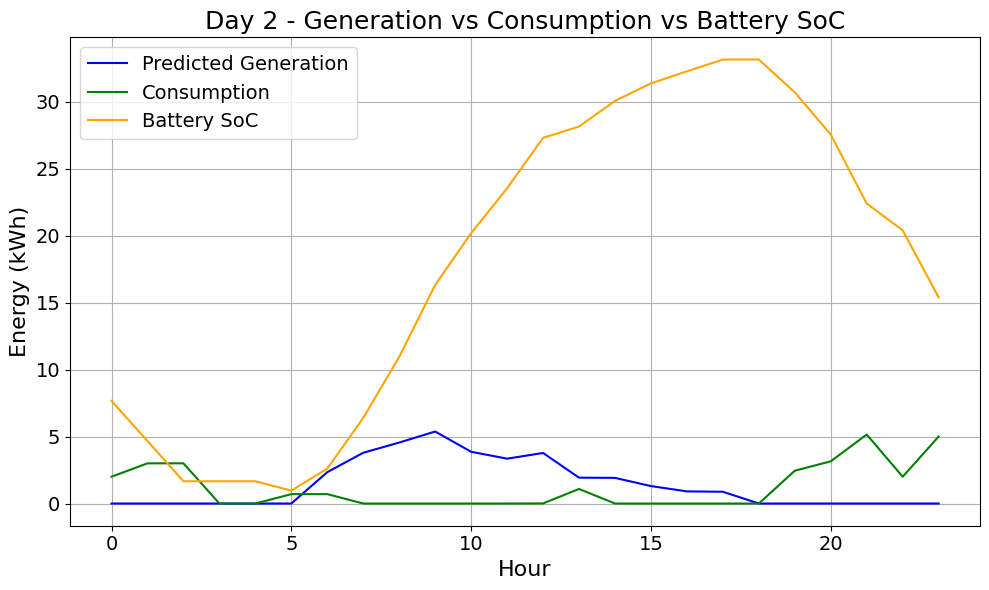}
        \caption{Day 2}
        \label{fig:day2_hourly_tracking}
    \end{subfigure}
    \caption{Hourly energy consumption, PV generation, and battery SoC for Day~1 and Day~2.}
    \label{fig:two_days_hourly_tracking}
\end{figure}

Figures~\ref{fig:day1_schedule} and \ref{fig:day2_schedule} present the optimized schedules, demonstrating the spillover of the washing task from Day~1 to Day~2. The impact of this scheduling on system operation is further illustrated in Figure~\ref{fig:two_days_hourly_tracking}, which depicts the hourly energy consumption, PV generation, and battery SoC across the two days, and shows how the battery SoC is carried over from one day to the next. As shown in Figs.~\ref{fig:day1_hourly_tracking} and \ref{fig:day2_hourly_tracking}, the battery state at 23:00 on Day 1 is 9.66, and at the start of Day 2 it decreases to 7.66 to maintain continuity, due to spillover consumption from the washing machine of 2 kW per hour.

%Representative SA appliance schedules for consecutive days are shown in Figures~6 and~7 for Day~1 and Day~2, highlighting spillover handling for the dishwasher and schedule continuity.
%. Hourly energy tracking in Figures~\ref{fig:day1_hourly_tracking} and \ref{fig:day2_hourly_tracking} illustrates appliance consumption, PV generation, and battery SoC, confirming adherence to operational constraints and efficient energy management.

\begin{figure}
    \centering
    \includegraphics[width=1\linewidth]{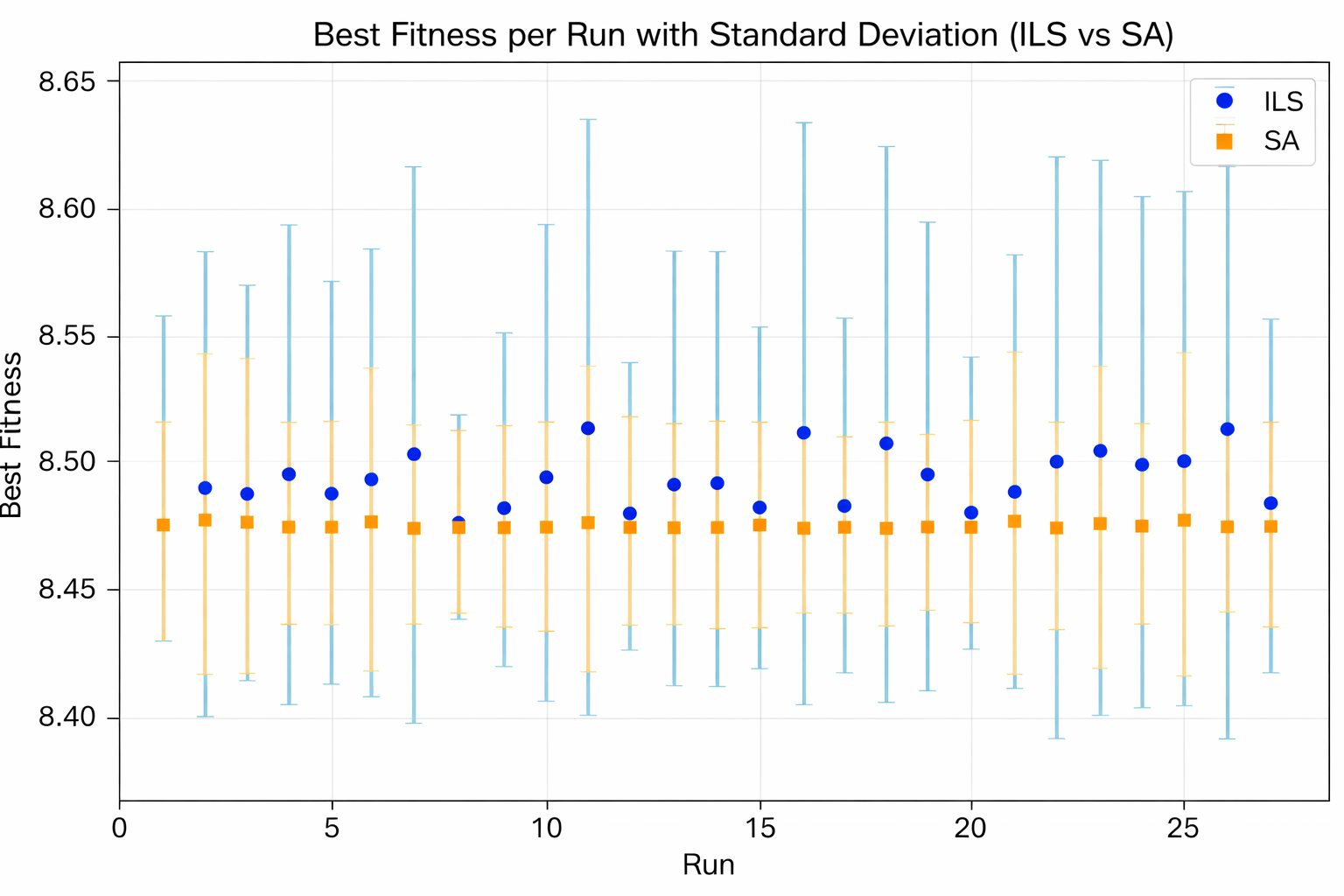}
   \caption{User dissatisfaction for SA and ILS over 30 runs.}
    \label{fig:user_dissatisfaction}
\end{figure}

After parameter tuning (Table~\ref{tab:tuned_hyperparams} for SA and ILS, respectively), the performance of SA and ILS was evaluated over 30 independent runs, each simulating 86 consecutive days. Figure~\ref{fig:user_dissatisfaction} shows the distribution of user dissatisfaction costs across these runs. SA consistently achieves lower mean fitness values with smaller variance in each run, whereas ILS exhibits higher variability, indicating less stable performance.

\begin{table}
%\caption{Best median fitness over 86 day schedules found by SA for various inverter sizes.}
\caption{Best median fitness over 86-day schedules for different inverter sizes (SA).}

\label{tab:first_run_inverter}
\begin{tabular}{ccrrrr}
\toprule
Inverter & Run & \multicolumn{4}{c}{Best Fitness} \\
\cmidrule(lr){3-6}
 &  & mean & std & min & max \\
\midrule
0.0 & 6.0 & 11000.000 & 0.000 & 11000.000 & 11000.000 \\
1.0 & 6.0 & 7002.925 & 0.019 & 7002.923 & 7003.102 \\
2.0 & 7.0 & 4006.455 & 10.686 & 4005.247 & 4104.400 \\
3.0 & 7.0 & 8.860 & 0.157 & 8.597 & 9.393 \\
4.0 & 6.0 & 8.662 & 0.151 & 8.517 & 9.293 \\
5.0 & 1.0 & 8.565 & 0.086 & 8.498 & 8.944 \\
7.5 & 3.0 & 8.465 & 0.047 & 8.460 & 8.897 \\
10.0 & 4.0 & 8.465 & 0.047 & 8.460 & 8.897 \\
\bottomrule
\end{tabular}
\end{table}

%\begin{table*}[htbp]
%\centering
%\caption{SA best fitness Grouped Summary by Inverter.}
%\label{tab:first_run_inverter}
%\begin{tabular}{cccccc}
%\toprule
%Inverter & Run & Best\_Fitness\_mean & Best\_Fitness\_std & Best\_Fitness\_min & Best\_Fitness\_max \\

%\midrule
%0.0 & 6.0 & 11000.000 & 0.000 & 11000.000 & 11000.000 \\
%1.0 & 6.0 & 7002.925 & 0.019 & 7002.923 & 7003.102 \\
%2.0 & 7.0 & 4006.455 & 10.686 & 4005.247 & 4104.400 \\
%3.0 & 7.0 & 8.860 & 0.157 & 8.597 & 9.393 \\
%4.0 & 6.0 & 8.662 & 0.151 & 8.517 & 9.293 \\
%5.0 & 1.0 & 8.565 & 0.086 & 8.498 & 8.944 \\
%7.5 & 3.0 & 8.465 & 0.047 & 8.460 & 8.897 \\
%10.0 & 4.0 & 8.465 & 0.047 & 8.460 & 8.897 \\
%\bottomrule
%\end{tabular}
%\end{table*}

Figure~\ref{fig:user_dissatisfaction} illustrates the distribution of user dissatisfaction costs across the 30 runs. SA consistently achieves lower mean fitness values with reduced variance, while ILS exhibits higher variability, indicating less stable performance.

\begin{comment}
\begin{figure}[htbp]
    \centering
    \includegraphics[width=\columnwidth]{30run_ils_sa.png}
    \caption{User dissatisfaction for SA and ILS over 30 runs.}
    \label{fig:user_dissatisfaction}
\end{figure}
\end{comment}
\subsubsection{Statistical Analysis}

To determine whether the observed performance differences between SA and ILS are statistically significant, the following hypotheses were formulated:
\begin{itemize}
    \item \textbf{Null hypothesis ($H_0$):} There is no significant difference between the fitness results obtained by SA and ILS.
    \item \textbf{Alternative hypothesis ($H_1$):} There is a significant difference between the fitness results obtained by SA and ILS.
\end{itemize}

The Shapiro--Wilk test was first applied to assess the normality of the fitness distributions, as summarized in Table~\ref{tab:shapiro_test}. Since the SA fitness distribution violated the normality assumption, a non-parametric Wilcoxon signed-rank test was used and confirmed a statistically significant difference between the two methods ($W = 0$, $p < 0.001$). To quantify the magnitude of this difference, effect size measures were computed. The rank-biserial correlation yielded $r_{rb} = 1.0$, and Cohen's effect size was $d = 2.55$, both indicating a large and practically meaningful performance advantage of SA over ILS. confirming that Simulated Annealing achieved consistently and substantially better fitness values than ILS.

\begin{table}[htbp]
\centering
\caption{Shapiro--Wilk test for fitness distribution normality}
\label{tab:shapiro_test}
\begin{tabular}{lccc}
\toprule
Algorithm & Statistic & $p$-value & Interpretation \\
\midrule
ILS & 0.9605 & 0.319 & Normal ($p > 0.05$) \\
SA  & 0.7065 & $1.9 \times 10^{-6}$ & Non-normal ($p < 0.05$) \\
\bottomrule
\end{tabular}
\end{table}

\subsubsection{Analysis Inverter Size}

In practical appliance scheduling, inverter capacity is an important constraint, as it limits how much power can be used at once. To study its effect, we ran a second set of experiments varying the inverter size while keeping everything else the same. Table~\ref{tab:first_run_inverter} shows the median daily best fitness values statistic for each inverter size, calculated from 10 independent runs. 
. The best scheduling performance was observed with an inverter size of 7.5, while below a certain threshold, meeting user constraints becomes difficult. These results highlight the link between inverter capacity and user satisfaction, emphasizing the importance of properly sizing the inverter. Moreover, as energy accumulates in the battery, multi-user sharing strategies could improve utilization and reduce costs. Further investigation of these trade-offs will be addressed in future work.

The results indicate that scheduling performance improves with larger inverter sizes, with the best performance observed at an inverter size of 7.5. Below a certain capacity, meeting user constraints becomes difficult, highlighting the importance of proper sizing of inverters.

%Inverter capacity plays a critical role in scheduling feasibility, as it is one of the key constraints in practical appliance scheduling. Table~\ref{tab:first_run_inverter} summarizes the statistics of the median ranked daily best fitness values for each inverter size, calculated from 10 independent runs: the mean, std. dev., minimum, and maximum fitness. The best SA performance was observed at an inverter size of 7.5, while below a certain threshold, meeting user constraints becomes challenging. These results highlight a correlation between inverter capacity and user satisfaction, emphasizing the importance of equipment sizing including batteries and PV cells to ensure feasible and effective appliance scheduling. Furthermore, as energy accumulates in the battery, these findings motivate exploring multi-user sharing strategies to optimize utilization and reduce costs. Investigation of these trade-offs will be further explored in future work.

\section{Conclusion}

This paper presented a metaheuristic-based appliance scheduling framework for residential energy management in solar-powered smart homes. The model enables appliance scheduling while considering continuous appliance durations, inverter and battery constraints, and multi-day task spillover. By integrating solar energy generation forecasts into metaheuristic optimization and sequential multi-day scheduling, the framework ensures adaptive scheduling aligned with renewable energy availability and user preferences, maintaining operational feasibility. 
Experimental evaluation over an 86 day simulation demonstrated that SA consistently outperforms ILS, achieving lower user dissatisfaction, more stable convergence, and improved utilization of renewable energy. Inverter sizing was shown to be critical for feasible scheduling, and the results indicate that high user satisfaction depends on selecting an appropriately sized inverter. Future work will extend the framework to multi-user environments and explore adaptive preference modeling to further enhance fairness, user satisfaction, and renewable energy utilization.

%These findings demonstrate the potential of integrating renewable generation forecasts directly into metaheuristic scheduling as a practical and effective solution for efficient demand-side energy management in renewable-powered smart homes, promoting sustainability without compromising  user comfort. F

\begin{acks}
ChatGPT 5.2 was used to refine grammar and flow of the text. The authors thoroughly reviewed all text prior to submission.
\end{acks}

\bibliographystyle{ACM-Reference-Format}
\bibliography{main}


\begin{thebibliography}{15}

%%% the Web address.

\ifx \showCODEN    \undefined \def \showCODEN     #1{\unskip}     \fi
\ifx \showDOI      \undefined \def \showDOI       #1{#1}\fi
\ifx \showISBNx    \undefined \def \showISBNx     #1{\unskip}     \fi
\ifx \showISBNxiii \undefined \def \showISBNxiii  #1{\unskip}     \fi
\ifx \showISSN     \undefined \def \showISSN      #1{\unskip}     \fi
\ifx \showLCCN     \undefined \def \showLCCN      #1{\unskip}     \fi
\ifx \shownote     \undefined \def \shownote      #1{#1}          \fi
\ifx \showarticletitle \undefined \def \showarticletitle #1{#1}   \fi
\ifx \showURL      \undefined \def \showURL       {\relax}        \fi
% invisible to TeX
\providecommand\bibfield[2]{#2}
\providecommand\bibinfo[2]{#2}
\providecommand\natexlab[1]{#1}
\providecommand\showeprint[2][]{arXiv:#2}

\bibitem[\protect\citeauthoryear{Bastianetto, Ceschia, and Schaerf}{Bastianetto et~al\mbox{.}}{2020}]%
        {bastianetto2020home}
\bibfield{author}{\bibinfo{person}{Edoardo Bastianetto}, \bibinfo{person}{Sara Ceschia}, {and} \bibinfo{person}{Andrea Schaerf}.} \bibinfo{year}{2020}\natexlab{}.
\newblock \showarticletitle{Solving a Home Energy Management Problem by Simulated Annealing}.
\newblock \bibinfo{journal}{\emph{Optimization Letters}} \bibinfo{volume}{15}, \bibinfo{number}{5} (\bibinfo{year}{2020}), \bibinfo{pages}{1553--1564}.
\newblock


\bibitem[\protect\citeauthoryear{Congram, Potts, and van~de Velde}{Congram et~al\mbox{.}}{2002}]%
        {congram2002iterated}
\bibfield{author}{\bibinfo{person}{R.~K. Congram}, \bibinfo{person}{C.~N. Potts}, {and} \bibinfo{person}{S. van~de Velde}.} \bibinfo{year}{2002}\natexlab{}.
\newblock \showarticletitle{An iterated dynasearch algorithm for the single-machine total weighted tardiness scheduling problem}.
\newblock \bibinfo{journal}{\emph{INFORMS Journal on Computing}} \bibinfo{volume}{14}, \bibinfo{number}{1} (\bibinfo{year}{2002}), \bibinfo{pages}{52--67}.
\newblock


\bibitem[\protect\citeauthoryear{Emden-Weiner and Proksch}{Emden-Weiner and Proksch}{1999}]%
        {emden-weiner1999best}
\bibfield{author}{\bibinfo{person}{T. Emden-Weiner} {and} \bibinfo{person}{M. Proksch}.} \bibinfo{year}{1999}\natexlab{}.
\newblock \showarticletitle{Best practice simulated annealing for the airline crew scheduling problem}.
\newblock \bibinfo{journal}{\emph{Journal of Heuristics}} \bibinfo{volume}{5}, \bibinfo{number}{4} (\bibinfo{year}{1999}), \bibinfo{pages}{419--436}.
\newblock


\bibitem[\protect\citeauthoryear{Gielen, Boshell, Saygin, Bazilian, Wagner, and Gorini}{Gielen et~al\mbox{.}}{2019}]%
        {gielen2019role}
\bibfield{author}{\bibinfo{person}{Dolf Gielen}, \bibinfo{person}{Francisco Boshell}, \bibinfo{person}{Deger Saygin}, \bibinfo{person}{Morgan~D Bazilian}, \bibinfo{person}{Nicholas Wagner}, {and} \bibinfo{person}{Ricardo Gorini}.} \bibinfo{year}{2019}\natexlab{}.
\newblock \showarticletitle{The role of renewable energy in the global energy transformation}.
\newblock \bibinfo{journal}{\emph{Energy Strategy Reviews}}  \bibinfo{volume}{24} (\bibinfo{year}{2019}), \bibinfo{pages}{38--50}.
\newblock


\bibitem[\protect\citeauthoryear{Hanafi and Kozan}{Hanafi and Kozan}{2014}]%
        {hanafi2014hybrid}
\bibfield{author}{\bibinfo{person}{R. Hanafi} {and} \bibinfo{person}{E. Kozan}.} \bibinfo{year}{2014}\natexlab{}.
\newblock \showarticletitle{A hybrid constructive heuristic and simulated annealing for railway crew scheduling}.
\newblock \bibinfo{journal}{\emph{Computers \& Industrial Engineering}}  \bibinfo{volume}{70} (\bibinfo{year}{2014}), \bibinfo{pages}{11--19}.
\newblock


\bibitem[\protect\citeauthoryear{Imran, Hafeez, Khan, Usman, Shafiq, Qazi, Khalid, and Thoben}{Imran et~al\mbox{.}}{2020}]%
        {imran2020heuristic}
\bibfield{author}{\bibinfo{person}{Adil Imran}, \bibinfo{person}{Ghulam Hafeez}, \bibinfo{person}{Imran Khan}, \bibinfo{person}{Muhammad Usman}, \bibinfo{person}{Zeeshan Shafiq}, \bibinfo{person}{Abdul~Baseer Qazi}, \bibinfo{person}{Azfar Khalid}, {and} \bibinfo{person}{Klaus-Dieter Thoben}.} \bibinfo{year}{2020}\natexlab{}.
\newblock \showarticletitle{Heuristic-based programmable controller for efficient energy management under renewable energy sources and energy storage system in smart grid}.
\newblock \bibinfo{journal}{\emph{IEEE Access}}  \bibinfo{volume}{8} (\bibinfo{year}{2020}), \bibinfo{pages}{139587--139608}.
\newblock


\bibitem[\protect\citeauthoryear{Kamyab}{Kamyab}{2025}]%
        {kamyab2025scheduling}
\bibfield{author}{\bibinfo{person}{Gholam-Reza Kamyab}.} \bibinfo{year}{2025}\natexlab{}.
\newblock \showarticletitle{Scheduling of home energy management systems for price-based demand response and end-users discomfort reduction}.
\newblock \bibinfo{journal}{\emph{Serbian Journal of Electrical Engineering}} \bibinfo{volume}{22}, \bibinfo{number}{1} (\bibinfo{year}{2025}), \bibinfo{pages}{17--34}.
\newblock


\bibitem[\protect\citeauthoryear{Kirkpatrick, Gelatt, and Vecchi}{Kirkpatrick et~al\mbox{.}}{1983}]%
        {kirkpatrick1983optimization}
\bibfield{author}{\bibinfo{person}{Scott Kirkpatrick}, \bibinfo{person}{C.~Daniel Gelatt}, {and} \bibinfo{person}{Mario~P. Vecchi}.} \bibinfo{year}{1983}\natexlab{}.
\newblock \showarticletitle{Optimization by simulated annealing}.
\newblock \bibinfo{journal}{\emph{Science}} \bibinfo{volume}{220}, \bibinfo{number}{4598} (\bibinfo{year}{1983}), \bibinfo{pages}{671--680}.
\newblock


\bibitem[\protect\citeauthoryear{Louren{\c{c}}o, Martin, and St{\"u}tzle}{Louren{\c{c}}o et~al\mbox{.}}{2003}]%
        {lourencco2003iterated}
\bibfield{author}{\bibinfo{person}{Helena~R Louren{\c{c}}o}, \bibinfo{person}{Olivier~C Martin}, {and} \bibinfo{person}{Thomas St{\"u}tzle}.} \bibinfo{year}{2003}\natexlab{}.
\newblock \showarticletitle{Iterated local search}.
\newblock In \bibinfo{booktitle}{\emph{Handbook of metaheuristics}}. \bibinfo{publisher}{Springer}, \bibinfo{pages}{320--353}.
\newblock


\bibitem[\protect\citeauthoryear{Matallanas, Castillo-Cagigal, Guti{\'e}rrez, Monasterio-Huelin, Caama{\~n}o-Mart{\'\i}n, Masa, and Jim{\'e}nez-Leube}{Matallanas et~al\mbox{.}}{2012}]%
        {matallanas2012neural}
\bibfield{author}{\bibinfo{person}{E. Matallanas}, \bibinfo{person}{Manuel Castillo-Cagigal}, \bibinfo{person}{A. Guti{\'e}rrez}, \bibinfo{person}{F. Monasterio-Huelin}, \bibinfo{person}{Estefan{\'\i}a Caama{\~n}o-Mart{\'\i}n}, \bibinfo{person}{D. Masa}, {and} \bibinfo{person}{J. Jim{\'e}nez-Leube}.} \bibinfo{year}{2012}\natexlab{}.
\newblock \showarticletitle{Neural network controller for active demand-side management with PV energy in the residential sector}.
\newblock \bibinfo{journal}{\emph{Applied Energy}} \bibinfo{volume}{91}, \bibinfo{number}{1} (\bibinfo{year}{2012}), \bibinfo{pages}{90--97}.
\newblock


\bibitem[\protect\citeauthoryear{Nakip, {\c{C}}opur, Biyik, and G{\"u}zelis}{Nakip et~al\mbox{.}}{2023}]%
        {nakip2023renewable}
\bibfield{author}{\bibinfo{person}{Mert Nakip}, \bibinfo{person}{Onur {\c{C}}opur}, \bibinfo{person}{Emrah Biyik}, {and} \bibinfo{person}{C{\"u}neyt G{\"u}zelis}.} \bibinfo{year}{2023}\natexlab{}.
\newblock \showarticletitle{Renewable energy management in smart home environment via forecast embedded scheduling based on Recurrent Trend Predictive Neural Network}.
\newblock \bibinfo{journal}{\emph{Applied Energy}}  \bibinfo{volume}{340} (\bibinfo{year}{2023}), \bibinfo{pages}{121014}.
\newblock


\bibitem[\protect\citeauthoryear{Shareef, Ahmed, Mohamed, and Al~Hassan}{Shareef et~al\mbox{.}}{2018}]%
        {shareef2018review}
\bibfield{author}{\bibinfo{person}{Hussain Shareef}, \bibinfo{person}{Maytham~S. Ahmed}, \bibinfo{person}{Azah Mohamed}, {and} \bibinfo{person}{Eslam Al~Hassan}.} \bibinfo{year}{2018}\natexlab{}.
\newblock \showarticletitle{Review on home energy management system considering demand responses, smart technologies, and intelligent controllers}.
\newblock \bibinfo{journal}{\emph{IEEE Access}}  \bibinfo{volume}{6} (\bibinfo{year}{2018}), \bibinfo{pages}{24498--24509}.
\newblock


\bibitem[\protect\citeauthoryear{St{\"u}tzle}{St{\"u}tzle}{1998}]%
        {stutzle1998ils}
\bibfield{author}{\bibinfo{person}{Thomas St{\"u}tzle}.} \bibinfo{year}{1998}\natexlab{}.
\newblock \bibinfo{booktitle}{\emph{Applying iterated local search to the permutation flow shop problem}}.
\newblock \bibinfo{type}{{T}echnical {R}eport} AIDA--98--04. \bibinfo{institution}{FG Intellektik, TU Darmstadt}.
\newblock


\bibitem[\protect\citeauthoryear{Varghese, Das, Ray, Mazid, Jahan, and Nur-E-Alam}{Varghese et~al\mbox{.}}{2025}]%
        {varghese2025optimisation}
\bibfield{author}{\bibinfo{person}{Benjohn~Koodakatt Varghese}, \bibinfo{person}{Narottam Das}, \bibinfo{person}{Biplob Ray}, \bibinfo{person}{Abdul~Md Mazid}, \bibinfo{person}{Israt Jahan}, {and} \bibinfo{person}{Mohamad Nur-E-Alam}.} \bibinfo{year}{2025}\natexlab{}.
\newblock \showarticletitle{Optimisation algorithms used in home energy management systems: A review}.
\newblock \bibinfo{journal}{\emph{Energy and Buildings}} (\bibinfo{year}{2025}), \bibinfo{pages}{116338}.
\newblock


\bibitem[\protect\citeauthoryear{Vilar and de~Mattos~Affonso}{Vilar and de~Mattos~Affonso}{2016}]%
        {vilar2016residential}
\bibfield{author}{\bibinfo{person}{Diego~Branches Vilar} {and} \bibinfo{person}{Carolina de Mattos~Affonso}.} \bibinfo{year}{2016}\natexlab{}.
\newblock \showarticletitle{Residential energy management system with photovoltaic generation using simulated annealing}. In \bibinfo{booktitle}{\emph{2016 13th International Conference on the European Energy Market (EEM)}}. IEEE, \bibinfo{pages}{1--6}.
\newblock


\end{thebibliography}

% \appendix



\end{document}